# Predictive State Representations: A New Theory for Modeling Dynamical Systems

**Satinder Singh**   **Michael R. James**   **Matthew R. Rudary**

Computer Science and Engineering
University of Michigan
Ann Arbor, MI 48109
{baveja,mrjames,mrudary}@umich.edu

## Abstract

Modeling dynamical systems, both for control purposes and to make predictions about their behavior, is ubiquitous in science and engineering. Predictive state representations (PSRs) are a recently introduced class of models for discrete-time dynamical systems. The key idea behind PSRs and the closely related OOMs (Jaeger's observable operator models) is to represent the state of the system as a set of predictions of observable outcomes of experiments one can do in the system. This makes PSRs rather different from history-based models such as $n^{th}$-order Markov models and hidden-state-based models such as HMMs and POMDPs. We introduce an interesting construct, the system-dynamics matrix, and show how PSRs can be derived simply from it. We also use this construct to show formally that PSRs are more general than both $n^{th}$-order Markov models and HMMs/POMDPs. Finally, we discuss the main difference between PSRs and OOMs and conclude with directions for future work.

## 1 Introduction

Modeling dynamical systems, both for control purposes and to make predictions about their behavior, is ubiquitous in science and engineering. Different disciplines often develop different mathematical formalisms for building models—differential equations in physics, finite-automata in computer science, logic and graphical models in artificial intelligence (AI), sequential decision processes in operations research (OR); these are but a few examples. Often the differences among the formalisms are motivated by properties of the class of dynamical systems of interest to the different fields. In this paper, we are exclusively interested in the class of general discrete-time, finite-observation, and stochastic dynamical systems. This class includes many of the problems of interest to the AI subfields of machine learning, reinforcement learning (RL), and planning. At the heart of current research in these subfields are hidden Markov models (HMMs) and their controlled counterparts partially observable Markov decision processes (POMDPs). Recently, Littman, Sutton, and Singh (2001) proposed a class of models called *predictive state representations* (PSRs) as an alternative to HMMs and POMDPs.

The key idea behind PSRs, and the closely related observable operator models or OOMs (Jaeger, 1997), is to represent the state of the system as a set of predictions of observable outcomes of tests or experiments one could do in the system. Thus, unlike hidden-state-based POMDP models, PSRs are expressed entirely in terms of observable quantities. Learning PSR models of dynamical systems from observation data should therefore be easier and less prone to local minima problems than learning POMDP models from observation data (Shatkay & Kaelbling, 1997). At the same time, PSRs do not have the severe limitations of history-based $n^{th}$-order Markov models, another class of models of dynamical systems based on purely observable quantities. Recent work on PSRs has begun to theoretically and empirically explore these advantages (Singh, Littman, Jong, Pardoe, & Stone, 2003; James & Singh, 2004).

In this paper we present a new and more comprehensive theory of PSRs than was available heretofore. The original development of PSRs by Littman et al. (2001) focused on their relationship to POMDPs and in particular showed how to convert a POMDP model to a PSR model. Here we present a new mathematical construct, the system-dynamics matrix ($\mathcal{D}$), that can be used to describe any controlled or uncontrolled dynamical system. This matrix $\mathcal{D}$ is not a model of the system but should be viewed as the system itself. We define the *linear dimension* of a dynamical sys-tem as the rank of its system-dynamics matrix. We use $\mathcal{D}$ to re-derive predictive state representations in a more general way than the derivation in Littman et al. (2001). We prove that dynamical systems with linear dimension $n$ can always be modeled by PSRs of size $n$ but that there exist such systems that cannot be modeled by any finite HMM/POMDP and any finite-order Markov model. Finally, we discuss the relationship between PSRs and the earlier work on OOMs by Jaeger (1997).

## 2 The System-Dynamics Matrix

An uncontrolled dynamical system can be viewed abstractly as a generator of observations. At time step $i$, it produces an observation $o_i$ from some set $\mathcal{O}$. The system itself can be viewed as a probability distribution over all possible futures of all lengths. A future is just a sequence of observations from the beginning of time. The prediction of a length-$k$ future $t = o^1 o^2 \ldots o^k$, denoted $p(t)$, is the probability that the first $k$ observations are precisely $t$, i.e., $p(t) = prob(o_1 = o^1, \ldots, o_k = o^k)$ where $o_i$ is the actual observation at time step $i$. A controlled dynamical system, on the other hand, takes inputs from some set $\mathcal{A}$ and generates observations from set $\mathcal{O}$. Thus a future in a controlled system is a sequence of action-observation pairs from the beginning of time. Again the system itself can be viewed as a probability distribution over all possible futures, but in this case conditional on the actions input to the system. Accordingly, a prediction for a length-$k$ future $t = a^1 o^1 \cdots a^k o^k$ is the probability that the first $k$ observations are $o^1 \cdots o^k$ given that the first $k$ actions are $a^1 \cdots a^k$, i.e., $p(t) = prob(o_1 = o^1, \ldots o_k = o^k | a_1 = a^1, \ldots a_k = a^k)$, where $a_i$ is the actual action at time step $i$. In controlled systems it is convenient to think of futures as *tests* or experiments one can do on the system. Thus, for test $t$ the prediction $p(t)$ is the probability of that test succeeding, i.e., of observing $t$'s sequence of observations upon doing $t$'s sequence of actions. Hereafter, we will refer to futures as tests for both controlled and uncontrolled systems.

Given an ordering over all possible tests $t_1 t_2 \ldots$, the system's probability distribution over all tests, defines an infinite *system-dynamics vector* $d$, such that the $i^{th}$ element of $d$ is the prediction of the $i^{th}$ test in the ordering, i.e., $d_i = p(t_i)$. Throughout, we will assume that the tests are arranged in order of increasing length and within the same length in lexicographic order. Figure 1a presents a pictorial view of the vector $d$. The predictions in $d$ have the following properties, illustrated in Figure 1b:

- $\forall i \; 0 \leq d_i \leq 1$

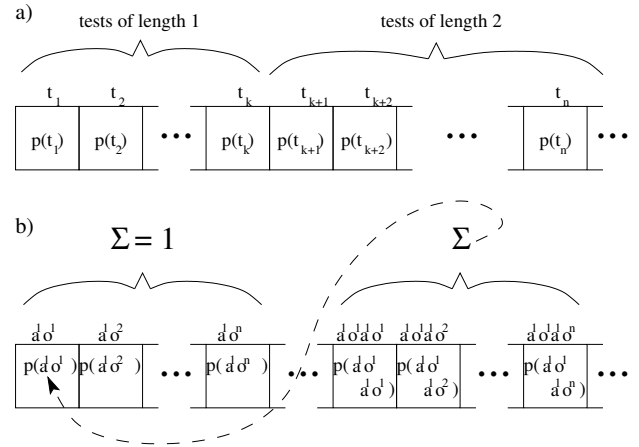

Figure 1: a) Each of $d$'s entries corresponds to the prediction of a test. b) Properties of the predictions imply structure in $d$.

- Let $T(\bar{a})$ be the set of tests whose action sequences equal $\bar{a}$. Then $\forall k \, \forall \bar{a} \in \mathcal{A}^k, \; \sum_{t \in T(\bar{a})} p(t) = 1$

- $\forall t \, \forall a \in \mathcal{A}, \; p(t) = \sum_{o \in \mathcal{O}} p(tao)$.

These properties imply that the infinite system-dynamics vector $d$ has a good deal of structure. One way to make this structure explicit is to consider a matrix, $\mathcal{D}$, whose columns correspond to tests and whose rows correspond to histories (see Figure 2). In an uncontrolled system a *history* is the sequence of observations from the beginning of time while in a controlled system a history is the sequence of action-observation pairs from the beginning of time. The interpretation is that a history is a future or test that has already happened. We define a history-conditional prediction for a test $t = a^1 o^1 \cdots a^k o^k$ given a history $h = a_1 o_1 \cdots a_j o_j$, as $p(t|h) = prob(o_{j+1} = o^1, \ldots, o_{j+k} = o^k | h, a_{j+1} = a^1, \ldots, a_{j+k} = a^k)$. The history-conditional prediction in the uncontrolled case can be defined analogously. We define an ordering over histories $h_1 h_2 \ldots$ similar to the ordering over tests, though we include the zero-length or *initial* history, $\phi$, as the first history in the ordering. Then

$$\mathcal{D}_{ij} = p(t_j | h_i) = \frac{p(h_i t_j)}{p(h_i)}, \qquad (1)$$

and each row of $\mathcal{D}$ has the same properties as $d$ had. In fact, the system-dynamics vector $d$ is the first row of the matrix $\mathcal{D}$ (see Figure 2). The matrix $\mathcal{D}$ with its infinitely many rows and columns is uniquely determined by the vector $d$ because both the numerator and the denominator of the rightmost term in Equation 1 are elements of $d$. We call $\mathcal{D}$ the *system-dynamics matrix*.

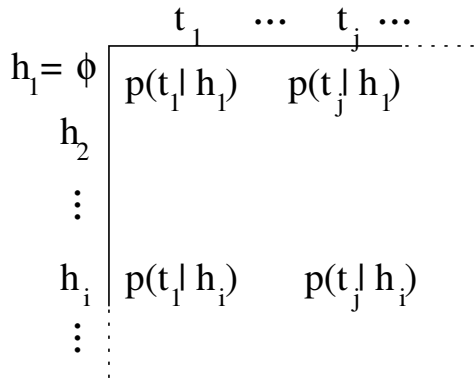

Figure 2: The rows in the system-dynamics matrix correspond to all possible histories (pasts), while the columns correspond to all possible tests (futures). The entries in the matrix are the predictions of tests given histories.

Because uncontrolled systems can be viewed as special cases of controlled systems in which there is only one action available, we will develop and present our results below for the controlled case. All of our results apply to the uncontrolled case unless noted otherwise.

## 3 Models

We emphasize that the system-dynamics matrix is not a model of the system; rather it is the system itself. Indeed, we will use $\mathcal{D}$ and dynamical system interchangeably. In other words, $\mathcal{D}$ places no constraints on the dynamical system at all other than our starting assumptions of discrete-time and finite actions and observations; for example, there is no assumption of stationarity of any sort. The matrix $\mathcal{D}$ forms the basis of our new theory of PSRs.

We begin by defining the concept of the *linear dimension* of a dynamical system as the *rank* of its corresponding system-dynamics matrix. The linear dimension of a system is equal to the dimension of the system as defined by Jaeger (1998) and, in the uncontrolled case, the minimum effective degree of freedom as defined by Ito, Amari, and Kobayashi (1992). We will show why the rank of $\mathcal{D}$ corresponds to a linear rather than nonlinear dimension later in this paper. But first, we remark that the rank of $\mathcal{D}$ is a measure of the complexity of the dynamical system and thus we should expect that a model of the dynamical system should have a complexity that is a function of the linear dimension.

Traditionally, a model of a dynamical system is something that can be used both as a simulator—to generate sequences of observations given sequences of actions—and as a predictor—to maintain state and make predictions about future behaviors while interacting with a system. In our context, an equivalent definition is that a model is something that can generate the system-dynamics matrix exactly. Most models are composed of the following pieces: a state representation that is a *sufficient statistic* of history, a specification of the initial state, and model parameters and update function that together define how the model updates the state as actions get taken and observations noted.

Before we derive a PSR model of a dynamical system from $\mathcal{D}$, we consider the relationship between $n^{th}$-order Markov models and $\mathcal{D}$ and then between HMMs/POMDPs and $\mathcal{D}$.

### 3.1 History-Based Models

An $n^{th}$-order Markov model makes the assumption that the next observation probabilities are conditionally independent of the history given the last $n$ action-observation pairs. The state of such a system in history $h$, denoted $s(h)$, is represented by the length-$n$ suffix of $h$; the initial state is the suffix that is equivalent to the null history. There are $k = (|\mathcal{A}||\mathcal{O}|)^n$ possible states. The parameters of this model are the $k|\mathcal{A}||\mathcal{O}|$ observation probabilities, arranged into $|\mathcal{A}|$ matrices $\{O^a\}$, where $O^a_{ij} = prob(o_j|s(h_i), a)$. The state is maintained simply; when a new action-observation pair is observed, the state is updated to the length-$n$ suffix of the new history.

Generating $\mathcal{D}$ from this model is straightforward: using the chain rule, $p(a^1 o^1 \cdots a^n o^n|h)$ is given by computing $prob(o^n|ha^1 o^1 \cdots a^{n-1} o^{n-1} a^n) \cdots prob(o^1|ha^1)$. Each of the probabilities in the product is obtained directly from the parameter matrices $\{O^a\}$ by mapping histories to states by using length-$n$ suffixes.

But how complex a dynamical system can be modeled by a $n^{th}$-order Markov model? In other words, what is the maximal rank of $\mathcal{D}$ that can be generated by such a model?

**Theorem 1** *The dynamical system corresponding to an $n^{th}$-order Markov model cannot have linear dimension greater than $k = (|\mathcal{A}||\mathcal{O}|)^n$.*

**Proof** Given that an $n^{th}$-order Markov model cannot distinguish between any two histories that have the same length-$n$ suffix, it is clear that the $\mathcal{D}$ matrix generated by such a model cannot have more than $k = (|\mathcal{A}||\mathcal{O}|)^n$ unique rows, and thus has rank at most $k$. The rank and therefore the linear dimension of the $\mathcal{D}$ generated by such a model is at most $k$. □

Later, we will show that there are systems with finite linear dimension that cannot be represented by an $n^{th}$-order Markov model for any finite $n$. Intuitively, this is because a matrix with finite rank may have infinitely many distinct rows. Taken together, these results show the known fact that $n^{th}$-order Markov models are quite limited in scope.

### 3.2 Models with Hidden States

POMDPs/HMMs are models based on the notion of underlying hidden or nominal states and directly address the limitations of history-based models. They maintain state information by keeping track of the probabilities of being in each of these nominal states as a function of history. Thus, the state representation of a $k$-state POMDP is a $k \times 1$ belief state vector $b(h)$, where $b_i(h)$ is the probability that the system is in nominal state $i$ given that history $h$ has been observed. The initial state is $b(\phi)$. The parameters of this model are the transition probabilities of the underlying MDP (that describes the unseen dynamics of the nominal states) and the observation probabilities; that is, a set of $k \times k$ stochastic matrices $\{T^a\}$, where $T^a_{ij}$ is the probability of transitioning from nominal state $i$ to nominal state $j$, and a set of $k \times k$ diagonal matrices $\{O^{ao}\}$ where $O^{ao}_{ii}$ is the probability of observing $o$ while leaving state $i$ by means of action $a$. Note that $\forall a \sum_{o \in \mathcal{O}} O^{ao} = I$. The state is updated by computing

$$b^T(hao) = \frac{b^T(h)T^a O^{ao}}{b^T(h)T^a O^{ao}\mathbf{1}}, \quad (2)$$

where $\mathbf{1}$ is a $k \times 1$ vector of all 1's.

$\mathcal{D}$ can be generated from a POMDP by generating each test's prediction as follows:

$$p(a^1 o^1 \cdots a^n o^n | h) = b(h) T^{a^1} O^{a^1 o^1} \cdots T^{a^n} O^{a^n o^n} \mathbf{1}$$

Again, we ask what the linear dimension is of systems that can be perfectly modeled by a POMDP.

**Theorem 2** *A POMDP with $k$ nominal states cannot model a dynamical system with linear dimension greater than $k$.*

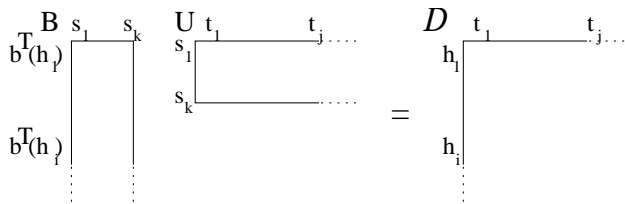

Figure 3: The system-dynamics matrix generated by a POMDP model with $k$ nominal states is the product of an $\infty \times k$ matrix ($B$) and a $k \times \infty$ matrix ($U$).

**Proof** Let $p(t|s_i)$ be the prediction for test $t$ generated by the POMDP for a belief state that assigns probability 1 to the system being in nominal state $s_i$ (note that it need not be possible to manipulate the system into such a belief state). Let $U$ be the $(k \times \infty)$ matrix whose $(ij)^{th}$ entry is $p(t_j|s_i)$. $U$ has $k$ rows, and so it has rank no more than $k$.

Now consider the $\mathcal{D}$ matrix generated by the POMDP. The row of $\mathcal{D}$ corresponding to a history $h$ is simply $b^T(h)U$, where $b(h)$ is the belief state after observing $h$. Thus $\mathcal{D}$ may be generated by computing the matrix multiplication $BU$, where $B$ is a $(\infty \times k)$ matrix with whose $i^{th}$ row is the belief state corresponding to history $h_i$ (see Figure 3). Because both $B$ and $U$ have rank no more than $k$, $\mathcal{D}$ has rank no more than $k$. Thus, no dynamical system that can be modeled by a POMDP with $k$ nominal states may have a linear dimension that is greater than $k$.
$\square$

Because an HMM is simply a POMDP with a single action, we can use Theorem 2 to assert:

**Corollary 3** *A HMM with $k$ nominal states cannot model a dynamical system with linear dimension greater than $k$.*

Although POMDPs are more expressive than $n^{th}$-order Markov models, there are dynamical systems with finite linear dimension that cannot be modeled by any finite POMDP. Jaeger (1998) presents an uncontrolled system that has a linear dimension of 3, but that cannot be modeled by any finite HMM. We do not repeat the construction here. And because all $n^{th}$-order Markov models can be represented as POMDPs, there are some dynamical systems with finite linear dimension that cannot be modeled by a finite POMDP or an $n^{th}$-order Markov model for any finite $n$.

## 4 Predictive State Representations

We now derive PSR models directly from the system-dynamics matrix. For any $\mathcal{D}$ with rank $k$, there must exist $k$ linearly independent columns and rows; these may not be unique. Consider any set of $k$ linearly independent columns of $\mathcal{D}$ and let the tests corresponding to those columns be $Q = \{q_1 \, q_2 \cdots q_k\}$. We call the tests in $Q$ the *core tests* (see Figure 4). Let the submatrix of $\mathcal{D}$ that contains just the columns for the core tests be denoted $\mathcal{D}(Q)$. The state representation of the PSR model is the set of predictions for the core tests. Thus, for any history $h$, the state of the PSR model is given by the column vector $p(Q|h) = [p(q_1|h) \, p(q_2|h) \cdots p(q_k|h)]$. The initial state is $p(Q|\phi)$, the entries of the first row in $\mathcal{D}(Q)$.

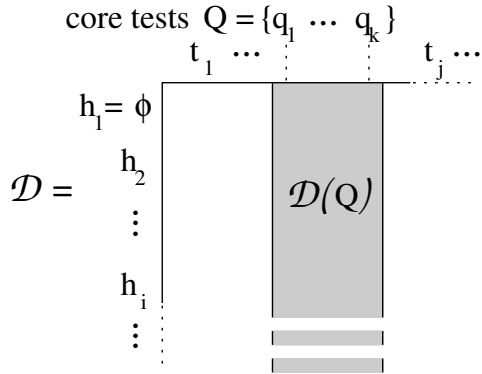

Figure 4: The core tests of a PSR are selected by finding $k$ linearly independent columns of $\mathcal{D}$.

Note that by the definition of rank of a matrix, all the columns of $\mathcal{D}$ are a linear combination of the columns in $\mathcal{D}(Q)$. In fact, for every test $t$, there exists a weight vector of length $k$, $m_t$, such that $\mathcal{D}(t)$, the column of $\mathcal{D}$ corresponding to test $t$, is given by $\mathcal{D}(t) = \mathcal{D}(Q)m_t$. This means that for any history $h$, $p(t|h) = p^T(Q|h)m_t$; thus this PSR is called a *linear* PSR. This allows us to compute a state update:

$$p(q_i|hao) = \frac{p(aoq_i|h)}{p(ao|h)} = \frac{p^T(Q|h)m_{aoq_i}}{p^T(Q|h)m_{ao}}$$

We can combine the entire update into a single equation by defining the matrices $M_{ao}$, where the $j^{th}$ column of $M_{ao}$ is simply $m_{aoq_j}$; then the update is given by

$$p^T(Q|hao) = \frac{p^T(Q|h)M_{ao}}{p^T(Q|h)m_{ao}}. \quad (3)$$

Therefore the model parameters in a PSR are $\{m_{aoq}\}_{a \in \mathcal{A}, o \in \mathcal{O}, q \in Q}$ and $\{m_{ao}\}_{a \in \mathcal{A}, o \in \mathcal{O}}$. Note that the model parameters may contain negative numbers in them; this sets them apart from POMDP model parameters.

Finally, a PSR model generates $\mathcal{D}$ as follows: for any test $t = a^1 o^1 \cdots a^n o^n$, its weight vector $m_t = M_{a^1 o^1} \cdots M_{a^{n-1} o^{n-1}} m_{a^n o^n}$ (Littman et al., 2001) can be computed from the model parameters and then used to generate the column $\mathcal{D}(t)$.

**Lemma 4** *A dynamical system with linear dimension $k$ can be modeled by a linear PSR with $k$ core tests.*

**Proof** This follows from the derivation of PSRs from $\mathcal{D}$ above. □

Because linear PSRs with $k$ core tests generate predictions for a test through a linear operation, they cannot represent systems with linear dimension more than $k$.

**Lemma 5** *A linear PSR with $k$ core tests cannot model a dynamical system with linear dimension greater than $k$.*

**Proof** In a PSR with $k$ core tests $Q$, the prediction for any test $t$ for a history $h$ is given by $p(t|h) = p^T(Q|h)m_t$ for some weight vector $m_t$. Thus in the $\mathcal{D}$ matrix generated by the PSR, the column corresponding to $t$, $\mathcal{D}(t)$, satisfies $\mathcal{D}(t) = \mathcal{D}(Q)m_t$. Thus, each column of $\mathcal{D}$ is a linear combination of the $k$ columns corresponding to the core tests, and $\mathcal{D}$ has rank no more than $k$. □

Thus, there is an equivalence between systems of finite linear dimension and systems that are modeled by linear PSRs with a finite number of core tests.

**Theorem 6** *Linear PSRs with $k$ core tests are equivalent to dynamical systems with linear dimension $k$.*

**Proof** This follows from Lemmas 4 and 5. □

Why is it that linear PSRs can model a larger class of dynamical systems than POMDPs? The main reason is that the update parameters of a PSR (see Equation 3) are **not** constrained to be non-negative while the update parameters of a POMDP (see Equation 2) are constrained to be non-negative.

## 5 Nonlinear Models

We defined the linear dimension of a dynamical system as the rank of the $\mathcal{D}$ matrix because, as we showed in deriving PSRs, there always exists a set of predictions of that size that allowed linear computation of the prediction for any test. In this sense the linear PSR state is a *linearly* sufficient statistic of the history. There may of course be a set of tests $N = \{n_1 \ldots n_c\}$ of size less than $|Q|$ whose predictions constitute a nonlinear sufficient statistic, such that for all tests $t$, $p(t|h) = f_t(p(N|h))$ for some nonlinear function $f_t$. Note that it is crucial that $f_t$ is independent of $h$, or else $p(N|h)$ won't be a sufficient statistic. When such a nonlinear sufficient statistic exists, a nonlinear PSR can be defined with update function

$$p(n_i|hao) = \frac{p(aon_i|h)}{p(ao|h)} = \frac{f_{aon_i}(p(Q|h))}{f_{ao}(p(Q|h))}.$$

It is helpful to see an example system in which a nonlinear PSR models a system more compactly than a linear PSR. Littman et al. (2001) created the float-reset problem, a POMDP with 5 states, illustrated in Figure 5. There are two actions and two observations. The float action moves to the state on the right or left with equal probability, and always results in observing 0. The reset action moves to the state on the far right.

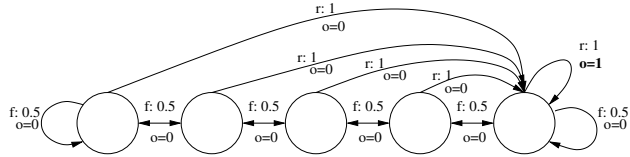

Figure 5: The float-reset problem.

If the system was already in that state, this results in an observation of 1; otherwise, the observation is 0. The system always starts in the far-right state.

Any linear PSR that models this system has (at least) 5 core tests. One such PSR has the core tests $\{r1, f0r1, f0f0r1, f0f0f0r1, f0f0f0f0r1\}$. Initially, and after a reset action, the predictions for these core tests are $[1, 0.5, 0.5, 0.375, 0.375]$. After a float action, each prediction is shifted left one place, and the last prediction is updated by $p(f0f0f0f0r1|hf0) = 0.0625p(r1|h) - 0.0625p(f0r1|h) - 0.75p(f0f0r1|h) + 0.75p(f0f0f0r1|h) + p(f0f0f0f0r1|h)$.

A glance at the initial prediction vector suggests there may be a pattern to the values that $p(r1|h)$ takes on. In fact, this is borne out by continuing the series; after the first transition (from 1 to 0.5), $p(r1|h)$ takes on the same value twice in a row as long as the float action is repeated. This twice-repeated value decreases monotonically and so $p(r1|h)$ and $p(f0r1|h)$ together uniquely determine the index into this infinite series. Therefore, one can always look up the new prediction for $p(f0r1|h)$ for a float action from the series. Even though we have not shown the explicit nonlinear calculation underlying the series, this argument suffices to show that the predictions for these two tests constitute a nonlinear sufficient statistic of the history. Thus there exists a nonlinear PSR for float-reset with the predictions for just two tests as state.

This can be taken further.

**Theorem 7** *There exist systems with nonlinear dimension exponentially smaller than their linear dimension.*

**Proof** (Sketch) Rudary and Singh (2004) present the rotate register system. This system is a $k$-bit rotate register, but only the far-left bit is observable. There are three actions: The register can be rotated either to the left or the right, and the visible bit can be "flipped" (i.e. changed from 0 to 1 or vice versa). A POMDP that models this system requires $2^k$ states, and a linear PSR requires $2^k$ core tests. However, a nonlinear PSR can model this system using only $k+1$ core tests. Thus, there is a nonlinear dimension to this system that is exponentially smaller than its linear dimension. See Rudary and Singh (2004) for details. □

## 6 PSRs and OOMs

So far we have shown that linear PSRs are more general than two currently popular models in AI, namely $n^{th}$-order Markov models and POMDPs. In fact, PSRs share many of their properties with Jaeger's OOMs. Jaeger (1998) has also developed many algorithms for OOMs and thus it is critical to understand the relationship between OOMs and PSRs so that research in PSRs can leverage existing work on OOMs correctly. We provide a beginning to this understanding in this paper.

Motivated by the fact that HMMs can be difficult to learn from observations of a system, Jaeger (1997) developed OOMs as an alternative model for uncontrolled dynamical systems.[1] Subsequently, he extended OOMs to Input/Output OOMs (IO-OOMs), which model controlled dynamical systems (Jaeger, 1998). In addition, Jaeger presented two versions each of OOMs and IO-OOMs: an uninterpretable version, in which the state vector has no interpretation, and an interpretable version, in which the elements of the state vector can be interpreted as predictions for a special kind of test in the system. Uninterpretable OOMs and IO-OOMs, while interesting, are not amenable to learning algorithms and thus we will focus on the interpretable versions of these models.

### 6.1 Interpretable OOMs for Uncontrolled Systems

The state vector in an interpretable OOM for an uncontrolled system is the set of predictions for a special set of tests. Consider all tests (as we defined them for uncontrolled systems) of some fixed length $m$ and partition them into $k$ subsets. Let each subset, so produced, be a union-test. The prediction of a union-test is the sum of the predictions for the tests in the union. Thus, for a given $m$ and $k$, the state vector for an OOM is a $k$-dimensional stochastic vector, i.e., for every history $h$ its state vector sums to one and each entry in the state vector is a probability.

On the one hand, union-tests are more general than the tests used in uncontrolled-PSRs because the former are unions of the latter tests. On the other hand, union-tests are less general than uncontrolled-PSR tests because they all have to be of the same length. In any case, OOM states are always stochastic vectors while PSR states have no such constraint. Given these differences what is the relationship between OOMs and uncontrolled PSRs?

---

[1]The notation used in this paper to describe OOMs departs somewhat from Jaeger's in order to remain consistent with the rest of the paper; instead of using $a$ for actions, he uses $r$, and instead of $o$ for observations, he uses $a$. In addition, we use the terms action and observation as opposed to Jaeger's terms input and output, and define interpretable OOMs and IO-OOMs in terms of tests instead of in terms of characteristic events.

**Theorem 8** *OOMs, both interpretable and uninterpretable, with dimension $k$ are equivalent to uncontrolled linear PSRs with $k$ core tests.*

**Proof** A linear PSR with $k$ core tests can model any system with linear dimension $\leq k$. Jaeger (1998) proved that $k$-dimensional interpretable OOMs are equivalent to $k$-dimensional uninterpretable OOMs, which in turn are equivalent to uncontrolled dynamical systems of linear dimension $k$. Thus, any system that can be modeled by an uncontrolled linear PSR with $k$ core tests can be modeled by an interpretable OOM of dimension $k$, and vice versa. □

So even though uncontrolled PSRs and interpretable OOMs have slightly different forms for uncontrolled dynamical systems, they are equivalent in power and in fact we have developed efficient algorithms (we omit these here for lack of space) for converting one to the other. The result for controlled dynamical systems is very different and we turn to that next.

### 6.2 Interpretable IO-OOMs for Controlled Systems

As with OOMs, the state in an interpretable IO-OOM is a vector of predictions for a special set of tests. Consider all tests (as we defined them for controlled systems) of some fixed length $m$ that share a particular action sequence $\bar{a}$ of length $m$. Partition this set of tests into $k$ union-tests. The prediction of a union-test is again the sum of the predictions for the tests in the union. Thus, for a given $m$, $\bar{a}$, and $k$, the state vector for an interpretable IO-OOM is a $k$-dimensional stochastic vector. The interesting constraint in interpretable IO-OOMs is the requirement that all union-tests share the same action sequence $\bar{a}$. This guarantees a stochastic vector as state but, as we will show below, places a crucial limit on the ability of interpretable IO-OOMs to model general dynamical systems.

**Theorem 9** *There exist controlled dynamical systems with finite linear dimension that cannot be modeled by any interpretable IO-OOM of any dimension.*

**Proof** We prove this theorem by presenting such a system. Figure 6 shows a POMDP with 4 nominal states that cannot be modeled by any interpretable IO-OOM. Recall (from Theorem 2) that any system that can be modeled by a POMDP with $k$ states has linear dimension $\leq k$, and so the POMDP in Figure 6 has a finite linear dimension.

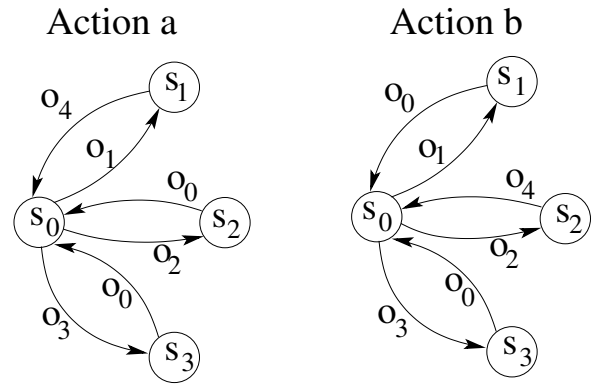

Figure 6: The POMDP representation of a system that cannot be correctly represented by an interpretable IO-OOM. The labels on the arcs are observations.

The key property of this system is that predicting the outcome of action $a$ will not differentiate between nominal states $s_2$ and $s_3$, while predicting the outcome of action $b$ will not differentiate between $s_1$ and $s_3$.

State $s_0$ is the initial nominal state of the system. Both actions produce the same transitions: When in nominal state $s_0$, the next nominal state is $s_1$, $s_2$, or $s_3$ with equal probability. When in one of the other nominal states, the next nominal state is $s_0$. When moving from $s_0$, both actions cause the same observations to be emitted: $o_i$ when entering $s_i$. However, when moving back to nominal state $s_0$, each action causes different observations. The observation emitted when moving to $s_0$ from $s_1$ using action $a$ is $o_4$ and $o_0$ otherwise; when action $b$ is executed, the observation is $o_4$ when leaving $s_2$ and $o_0$ otherwise.

It is easy to see that history provides enough information to determine the nominal state with certainty. At times $0, 2, 4, \ldots$, the system is in nominal state $s_0$; the observation emitted by the system identifies the next nominal state. Since the observation emitted when returning to $s_0$ is deterministic given the nominal state and action, any complete model of the system should predict perfectly the observations at times $1, 3, 5, \ldots$.

We now prove that no interpretable IO-OOM can model this system. We start by showing that no interpretable IO-OOM with an input sequence of length 1 can model the system and complete the proof by showing that no longer input sequence is sufficient.

Let us start with the input sequence $\bar{a} = a$. There are five observations, so there are five events that use this as their action sequence. None of these events has a different probability when starting in $s_2$ than it does when starting in $s_3$. Thus the state vector for an interpretable IO-OOM with input sequence $a$ would be the same when the system was in either of these states, and would be unable to predict the observation emitted when leaving those nominal states using action $b$.

Similarly, an interpretable IO-OOM with input sequence $\bar{a} = b$ would be unable to differentiate between $s_1$ and $s_3$ and would not correctly predict the observation emitted when leaving those nominal states using action $a$.

Furthermore, no longer sequence is sufficient. Because the system always returns to $s_0$ on even time steps, effectively resetting the system, knowing the behavior of the system two time steps or more in advance does not provide any information about the current state of the system. Thus, no interpretable IO-OOM can model this system. □

A corollary follows directly from Theorems 6 and 9:

**Corollary 10** *There exist controlled dynamical systems that can be modeled by PSRs that cannot be modeled by any interpretable IO-OOM; PSRs are more general than interpretable IO-OOMs.*

This severely limits the utility of IO-OOMs. Because uninterpretable IO-OOMs are not verifiable in the sense that they are not directly based on data produced by the system, it is difficult to infer them from such data. In fact, the learning algorithm that Jaeger (1998) presents for IO-OOMs only works for the interpretable version.

## 7 Conclusion and Future Work

We have introduced the system-dynamics matrix, a mathematical construct that provides an interesting way of looking at discrete dynamical systems. Using this matrix, we have re-derived PSRs in a very simple way and showed that they are strictly more general than both POMDPs and $n^{th}$-order Markov models. The original formulation of PSRs by Littman et al. (2001), though it resulted in exactly the same model as this derivation, was arrived at through POMDPs and was therefore more limited and complex. In addition, we have shown that there exist dynamical systems with nonlinear dimension that is exponentially smaller than the linear dimension of those systems. Finally, we have shown that in the case of uncontrolled dynamical systems, linear PSRs and interpretable OOMs are equivalent, while in the case of controlled dynamical systems, interpretable IO-OOMs are less general than linear PSRs.

Taken together our results form the beginnings of a theory of PSRs though, of course, much remains to be done. We are currently pursuing the development of general nonlinear models by attempting to estimate the nonlinear dimension directly from the system-dynamics matrix.

### Acknowledgements

The authors gratefully acknowledge many insightful discussions on the content of this paper and on PSRs more generally with Richard Sutton, Michael Littman and Britton Wolfe. The authors also thank the anonymous reviewers for their helpful comments.